\documentclass[default,iicol]{sn-jnl}

\usepackage{graphicx}%
\usepackage{multirow}%
\usepackage{amsmath,amssymb,amsfonts}%
\usepackage{amsthm}%
\usepackage{mathrsfs}%
\usepackage[title]{appendix}%
\usepackage{xcolor}%
\usepackage{textcomp}%
\usepackage{manyfoot}%
\usepackage{booktabs}%
\usepackage[lined,boxed,commentsnumbered]{algorithm2e}%
\usepackage{algpseudocode}%
\usepackage{listings}%

\usepackage{tabulary,multirow,array,float,colortbl,array}
\newcolumntype{I}{!{\vrule width 3pt}}
\newlength\savedwidth

\newlength\savewidth
\newcommand\shline{\noalign{\global\savewidth\arrayrulewidth
\global\arrayrulewidth 1pt}%
\hline
\noalign{\global\arrayrulewidth\savewidth}}

\newcolumntype{x}[1]{>{\centering\arraybackslash}p{#1pt}}

\newcommand{\tablestyle}[2]{\setlength{\tabcolsep}{#1}\renewcommand{\arraystretch}{#2}\centering\footnotesize}

\theoremstyle{thmstyleone}%
\theoremstyle{thmstyletwo}%

\theoremstyle{thmstylethree}%

\begin{document}

\title[Article Title]{AnomalyXFusion: Multi-modal Anomaly Synthesis with Diffusion}

\author[1]{\fnm{Jie} \sur{Hu}}\email{hujie.cpp@gmail.com}
\equalcont{These authors contributed equally to this work.}
\author[2]{\fnm{Yawen} \sur{Huang}}\email{yawenhuang@tencent.com}
\equalcont{These authors contributed equally to this work.}
\author[3]{\fnm{Yilin} \sur{Lu}}\email{yilinlu@stu.xmu.edu.cn}
\equalcont{These authors contributed equally to this work.}
\author*[1,4]{\fnm{Guoyang} \sur{Xie}}\email{guoyang.xie@ieee.org}
\author*[1]{\fnm{Guannan} \sur{Jiang}}\email{jianggn@catl.com}
\author[2]{\fnm{Yefeng} \sur{Zheng}}\email{yefeng.zheng@gmail.com}
\author[4]{\fnm{Zhichao} \sur{Lu}}\email{luzhichaocn@gmail.com}

\affil[1]{\orgdiv{Department of Intelligent Manufacturing}, \orgname{CATL}, \orgaddress{\city{Ningde}, \country{China}}}

\affil[2]{\orgdiv{Jarvis Research Center}, \orgname{Tencent Youtu Lab}, \orgaddress{ \city{Shenzhen}, \country{China}}}

\affil[3]{\orgdiv{School of Informatics}, \orgname{Xiamen University}, \orgaddress{\city{Xiamen}, \country{China}}}

\affil[4]{\orgdiv{Department of Computer Science}, \orgname{City University of Hong Kong}, \orgaddress{\city{Hongkong}, \country{China}}}


\abstract{
Anomaly synthesis is one of the effective methods to augment abnormal samples for training.
However, current anomaly synthesis methods predominantly rely on texture information as input, which limits the fidelity of synthesized abnormal samples.
Because texture information is insufficient to correctly depict the pattern of anomalies, especially for logical anomalies.
To surmount this obstacle, we present the AnomalyXFusion framework, designed to harness multi-modality information to enhance the quality of synthesized abnormal samples.
The AnomalyXFusion framework comprises two distinct yet synergistic modules: the Multi-modal In-Fusion (MIF) module and the Dynamic Dif-Fusion (DDF) module.
The MIF module refines modality alignment by aggregating and integrating various modality features into a unified embedding space, termed X-embedding, which includes image, text, and mask features.
Concurrently, the DDF module facilitates controlled generation through an adaptive adjustment of X-embedding conditioned on the diffusion steps.
In addition, to reveal the multi-modality representational power of AnomalyXFusion, we propose a new dataset, called MVTec Caption.
More precisely, MVTec Caption extends 2.2k accurate image-mask-text annotations for the MVTec AD and LOCO datasets.
Comprehensive evaluations demonstrate the effectiveness of AnomalyXFusion, especially regarding the fidelity and diversity for logical anomalies.
Project page: \url{https://github.com/hujiecpp/MVTec-Caption}
}


\maketitle

\section{Introduction}
Image Anomaly Detection (IAD) is of paramount importance in industrial manufacturing~\cite{liu2024deep,xie2024iad}.
Most of the IAD algorithms employ unsupervised learning methods that heavily rely on normal samples for training, or few-shot supervised learning methods.
Although these methods perform very well on anomaly detection, they have limited performance on localizing the anomalies.
To address this issue, researchers have begun to convert the anomaly localization problem into an anomaly segmentation problem~\cite{zavrtanik2021draem,yang2023memseg}.
Most anomaly segmentation models require a large number of abnormal samples for training.
However, acquiring a wide variety of defect samples can be challenging and resource-intensive, often leading to a scarcity of diverse abnormal data~\cite{bergmann2019mvtec,sindagi2017domain,diers2023survey}.
Hence, it is important and urgent to address the scarcity of abnormal samples~\cite{duan2023few,hu2024anomalydiffusion,lin2021few}.

Anomaly synthesis has emerged as a pivotal technique to mitigate the issue of limited abnormal sample availability~\cite{li2021cutpaste,zavrtanik2021draem,zhang2021defect,niu2020defect}.
It enables the creation of synthetic defect samples, thereby enriching the dataset and enhancing the robustness of detection models.
Despite the promise of anomaly synthesis, existing methods are predominantly single-modal, focusing on textural imagery, which limits the precision and variety of synthesized content.
The generation of complex defects, such as structural and logical anomalies, is particularly challenging with single-modal approaches, as they lack the capacity to capture the intricacies of the product's geometry and function.

\begin{figure}
\centering
\includegraphics[width=1.0\linewidth]{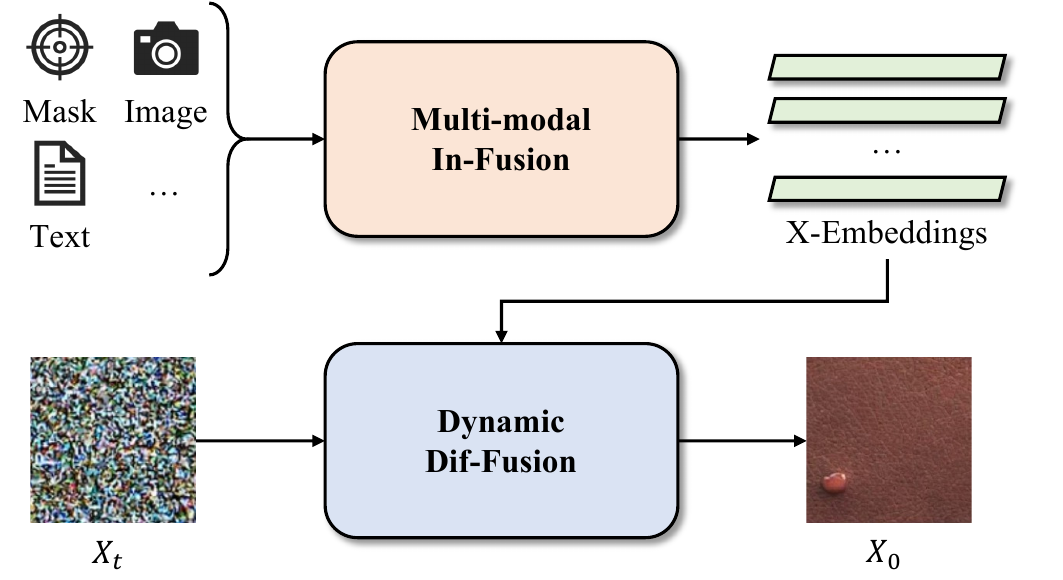} 
\vspace{-4.5mm}
\caption{
\textbf{Framework of AnomalyXFusion}. AnomalyXFusion aims to integrate multi-modal data into the process of anomaly synthesis.
Specifically, the Multi-modal In-Fusion (MIF) module initially fuses various data modalities to form X-embeddings.
Subsequently, the embeddings are dynamically adjusted based on the diffusion steps via the Dynamic Dif-Fusion (DDF) module.
}\label{fig1}
\end{figure}

\begin{figure*}
\centering
\includegraphics[width=1.0\linewidth]{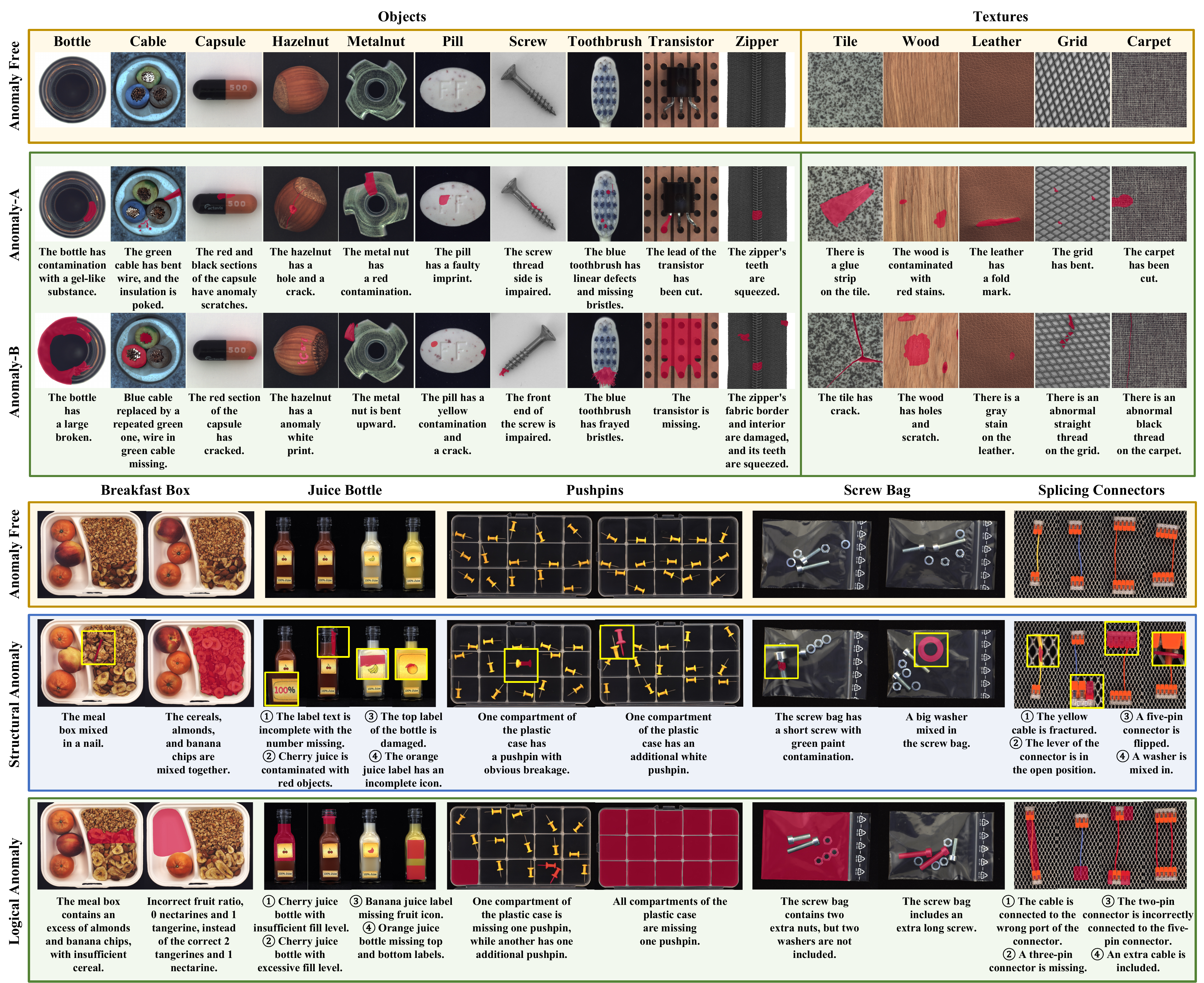} 
\vspace{-3.5mm}
\caption{
\textbf{MVTec Caption Dataset}.
We present the MVTec Caption dataset, an enhancement to the existing MVTec AD and LOCO datasets, which now includes precise textual annotations for anomalies.
These detailed descriptions not only resolve semantic ambiguities inherent in image-mask annotations but also facilitate the identification of logical anomalies.
For instance, in the cable category, the annotations can clarify color errors, determine whether wires are missing or damaged; in the breakfast box category, they can specify incorrect counts of fruits or the presence of mixed grains.
Without such precise semantic information, these distinctions would be challenging to ascertain.
}\label{fig2}
\end{figure*}

To address these limitations, this study introduces the AnomalyXFusion framework, a state-of-the-art multi-modal approach designed to synthesize anomalies within the context of industrial manufacturing.
The AnomalyXFusion framework aims to harness the diverse information present in industrial environments, thereby enhancing the generation of abnormal samples with greater authenticity and diversity.
As illustrated in Fig.~\ref{fig1}, the framework is comprised of two distinct yet synergistic modules: a Multi-modal In-Fusion (MIF) module and a Dynamic Dif-Fusion (DDF) module.
The MIF module enhances the process of aligning modalities by combining and incorporating various modal data, such as visual imagery, textual descriptions, binary masks, and depth metrics, into a unified embedding space referred to as X-Embeddings.
Simultaneously, the DDF module enables controlled generation by adaptively adjusting embedding weights based on the advancement of diffusion steps.
In addition, to address the existing gap in multi-modality anomaly detection datasets, characterized by the lack of detailed descriptions, we present the MVTec Caption dataset.
As depicted in Fig.~\ref{fig2}, this dataset encompasses 2.2k image-mask-text descriptions that are meticulously curated to provide an accurate portrayal of anomalies present within the images.
Empirical evaluations show that AnomalyXFusion markedly outperforms contemporary methods in both the authenticity and diversity of generated anomalies on the MVTec Caption dataset.
The framework significantly augments the efficacy of subsequent anomaly detection procedures, including advanced localization and segmentation tasks.
In short, our contributions can be encapsulated as follows:
\begin{itemize}
\item \textbf{AnomalyXFusion Framework}: We propose a novel framework, AnomalyXFusion, specifically tailored for enhancing the synthesis of anomalies in industrial manufacturing. This framework addresses the limitations imposed by the scarcity of abnormal samples and moves beyond the reliance on single-modal data, which is a common constraint in traditional anomaly synthesis methods.
\item \textbf{Multi-modal In-Fusion (MIF) Module}: The MIF module effectively integrates visual, textual, and mask data into a unified embedding space, enabling a more nuanced and comprehensive depiction of abnormalities.
\item \textbf{Dynamic Dif-Fusion (DDF) Module}: We present the DDF module, which introduces an adaptive mechanism for adjusting the weighting of embeddings based on the diffusion steps. This dynamic adjustment mechanism allows for greater control over the generation process, ensuring that the synthesized anomalies are not only diverse but also contextually relevant.
\item \textbf{MVTec Caption Dataset}: We revisit the MVTec AD and LOCO datasets, contributing 2.2k image-text pairs that offer precise descriptions of the anomalies. This augmentation is intended to enable a deeper and more nuanced understanding of the anomalies.
\item \textbf{Empirical Evaluation and Performance}: Through extensive empirical evaluations, we demonstrate that AnomalyXFusion significantly outperforms state-of-the-art methods in terms of generation authenticity and diversity. The enhanced synthesis capabilities of our framework have markedly improved anomaly detection, localization, and classification tasks.
\end{itemize}

\section{Related Work}

\subsection{Image Generation}
Image generation has been a vibrant area of research in computer vision and machine learning, with several notable models and techniques contributing to the field's advancement.
Variational Autoencoders (VAEs)~\cite{kingma2013auto} are generative models that combine the capabilities of autoencoders and Bayesian inference.
They consist of an encoder that maps the input data to a probability distribution and a decoder that samples from this distribution to generate new instances of data.
Generative Adversarial Networks (GANs)~\cite{goodfellow2014generative} consist of two neural networks, a generator and a discriminator, that are trained in opposition.
The generator creates images, and the discriminator evaluates them, with the goal of producing images that the discriminator cannot distinguish from real ones.
GANs have been incredibly successful in generating high-fidelity images and have led to various architectures like DCGAN~\cite{radford2015unsupervised}, BigGAN~\cite{brock2018large}, and StyleGAN~\cite{karras2019style}, each with its own improvements and applications.
Diffusion models are a class of generative models that simulate a diffusion process, gradually transforming a random noise distribution into a data distribution.
They are trained to reverse this process, learning to generate data by predicting how to remove noise from intermediate diffusion steps.
Denoising Diffusion Probabilistic Model (DDPM)~\cite{ho2020denoising} is a type of diffusion model that has gained significant attention for its ability to generate high-quality images.
It defines a forward diffusion process that progressively adds noise to data and a reverse process that gradually removes this noise.
The model learns this reverse process by minimizing the discrepancy between its predictions and the true data distribution.
DDPM has been noted for its remarkable fidelity and potential to synthesize complex data, such as images.
Latent Diffusion Model (LDM)~\cite{rombach2022high} is an advancement over DDPM that focuses on the latent space of the diffusion process.
Instead of working directly on the data space, LDM operates in a lower-dimensional latent space, which allows for more efficient sampling and improved image quality.
LDMs have demonstrated the ability to generate high-resolution images, making them a powerful tool for various image-generation tasks.
In this work, we integrate the LDM with a dynamic process that adaptively adjusts the input embeddings' weights for controlled generation, based on the diffusion steps.

\begin{figure}
\centering
\includegraphics[width=1.0\linewidth]{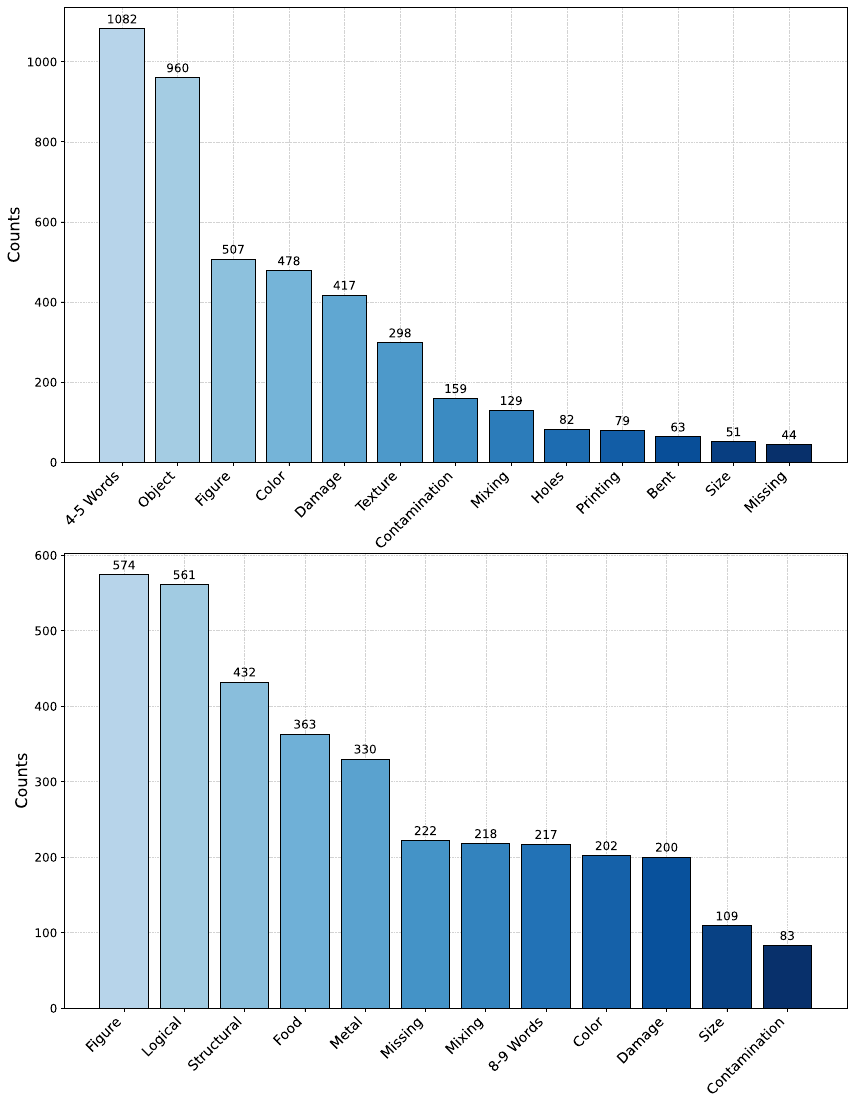} 
\vspace{-3.5mm}
\caption{
\textbf{Analysis of Attributions in the Proposed MVTec Caption Dataset}. 
As illustrated in the figure, the dataset encompasses a diverse array of textual keywords, which contributes to the diversity.
}\label{fig3}
\end{figure}

\subsection{Anomaly Synthesis}
Anomaly synthesis is a key technique, particularly for improving the performance of anomaly detection systems in scenarios where real defect data are limited.
In the literature, methods such as DRAEM~\cite{zavrtanik2021draem}, CutPaste~\cite{li2021cutpaste,niu2020defect}, and EasyNet~\cite{chen2023easynet} have been employed to augment normal samples with unrelated textures or pre-existing anomalies through a cropping and pasting approach.
However, these techniques have been criticized for yielding anomalies that lack realism or for exhibiting a constrained diversity in the generated samples.
Subsequently, GANs have been extensively applied for anomaly synthesis due to their ability to generate high-fidelity images.
GAN-based models such as SDGAN~\cite{niu2020defect, zhang2021defect} and Defect-GAN~\cite{niu2020defect} synthesize anomalies on normal samples through learning from defect data.
However, these models necessitate a substantial dataset of anomalies, and they are unable to produce corresponding anomaly masks.
DFMGAN~\cite{duan2023few} adapts a pre-trained StyleGAN2~\cite{viazovetskyi2020stylegan2} model to generate anomalies. Nevertheless, this approach falls short in terms of generation authenticity and precise alignment of the synthesized anomalies with the masks.
Recently, diffusion models have emerged as a promising avenue for anomaly synthesis, capitalizing on their capacity to produce a diverse array of anomalies with smooth transitions from noise to coherent data, thereby enhancing the realism of the synthesized defects.
AnomalyDiffusion~\cite{hu2024anomalydiffusion} represents an innovative approach within this domain, adept at crafting a spectrum of anomalies with high fidelity.
Specifically, AnomalyDiffusion integrates the strong prior information from pre-trained Latent Diffusion Models (LDMs), which allows for the generation of diverse and authentic anomalies even from a limited number of training samples.
However, prevailing methods for anomaly synthesis typically rely on a single modality, such as textual images, which limits their efficacy in generating complex defects, including structural and logical anomalies.
In this paper, we introduce the AnomalyXFusion framework, which enables the generation of defects based on multiple modalities, significantly enriching the diversity and authenticity of the synthesized anomalies.

\subsection{Anomaly Segmentation}
The anomaly segmentation task encompasses the interconnected objectives of anomaly detection, localization, and classification.
A range of methodologies has been developed, focusing on image reconstruction techniques.
For instance, the approach in~\cite{schlegl2017unsupervised} employs image reconstruction to identify anomalies by comparing discrepancies between reconstructed and actual images.
Similarly, f-AnoGAN \cite{schlegl2019f} leverages GANs to enable rapid anomaly detection, eliminating the need for labeled data.
The method described in~\cite{liang2023omni} also relies on reconstruction but emphasizes selecting optimal frequency channels to enhance anomaly detection accuracy.
In addition to reconstruction-based methods, deep feature modeling approaches have been proposed to create a comprehensive feature space for input images.
These approaches, such as the one described in CFA~\cite{lee2022cfa}, aim to identify anomalies by analyzing feature variations within this space..
The work in~\cite{cao2022informative} focuses on transferring knowledge from informative features to improve segmentation accuracy.
TRIAD~\cite{roth2022towards} addresses the challenge of detecting all possible types of anomalies in industrial settings, while MemKD~\cite{gu2023remembering} introduces a novel technique for guiding the anomaly detection process by remembering normal patterns.
In addition, M3DM~\cite{wang2023multimodal} explores the use of multiple data types to enhance the detection of anomalies in industrial applications.
Real3D-AD~\cite{liu2024real3d} discusses the possibility of using full-angle point clouds for anomaly detection.
Recent advancements in the field include the development of zero/few-shot learning techniques~\cite{xie2022pushing,jeong2023winclip,cao2023segment} for anomaly classification and segmentation, such as the method proposed in WinCLIP~\cite{jeong2023winclip}, which aims to classify and segment anomalies with limited or no training data.
The use of hybrid prompt regularization has also been explored as a means to segment any anomaly, thereby expanding the capabilities of anomaly detection systems.
These developments reflect the ongoing progress in the domain of anomaly perception, with a focus on improving the accuracy and efficiency of detection, localization, and classification tasks.
The integration of these innovative techniques into practical applications holds the potential to significantly enhance the performance of anomaly detection systems in various industries.
This paper presents a solution to the challenge of high-quality data scarcity in anomaly perception by effectively generating diverse synthetic anomalies from limited real data.
This approach enhances dataset robustness and improves model training, leading to better anomaly detection performance.

\section{MVTec Caption Dataset}
Existing anomaly detection datasets, despite offering abnormal samples with corresponding labels and location masks, frequently fall short of providing the nuanced semantic descriptions necessary for a precise characterization of anomalies.
This limitation is particularly pronounced when identifying and categorizing logical irregularities, as traditional annotations, including anomaly masks and image-level labels, prove insufficient.
Recent progress in language and segmentation models has prompted the exploration of textual prompts to enhance anomaly detection capabilities, as evidenced by studies such as WinCLIP~\cite{jeong2023winclip} and SAA~\cite{cao2023segment}.
Nevertheless, the lack of detailed descriptions for logical anomalies continues to hinder the effectiveness of these methods in detecting such anomalies.
For instance, in the cable category, annotations could clarify color errors, ascertain whether wires are missing or damaged; in the breakfast box category, they could specify incorrect counts of fruits or the presence of mixed grains.
The absence of precise semantic information renders these distinctions difficult to identify.
To address these limitations, we have meticulously annotated the MVTec AD~\cite{bergmann2021mvtec,bergmann2019mvtec} and MVTec LOCO~\cite{bergmann2022beyond} datasets with extensive captions. Specifically, we have provided 1258 and 982 captions for the AD and LOCO datasets, respectively.
As illustrated in Fig.~\ref{fig2}, the inclusion of detailed captions significantly facilitates the clear assessment of anomalies.
Moreover, the distribution of caption attributes, as shown in Fig.~\ref{fig3}, highlights the diversity and comprehensiveness of our proposed dataset, thereby enhancing its utility in anomaly detection research.

\section{Method}

\subsection{AnomalyXFusion}
As depicted in Fig.~\ref{fig1}, the AnomalyXFusion framework incorporates two pivotal components: the Multi-modal In-Fusion (MIF) module and the Dynamic Dif-Fusion (DDF) module.
The MIF module is responsible for the preliminary processing of multi-modal information, which includes extraction, alignment, and integration of different data types.
Within the text modality, the CLIP text encoder~\cite{radford2021learning} is utilized to extract features and generate semantic embeddings.
For the spatial (or mask) modality, which delineates the location of anomalies, a spatial mask encoder~\cite{hu2021istr,hu2024istr} is employed to produce location embeddings.
The textural information is processed through the CLIP image encoder~\cite{radford2021learning} to capture pertinent textural features. These features are then integrated into the textual embeddings for the subsequent training phase.
The distinct multi-modal embeddings, namely, semantic, location, and textural embeddings, are subsequently synthesized through a multi-modal aggregator, resulting in the creation of aggregated X-Embeddings.
These X-Embeddings are subsequently refined by the DDF module for conditional anomaly generation, with adaptive adjustments made throughout the dynamic diffusion process to correspond with the progression of the diffusion steps.

The subsequent sections will provide an in-depth exploration of the operational mechanisms underlying each module and their individual contributions to the anomaly synthesis process.

\subsection{Multi-modal In-Fusion}

\noindent\textbf{Semantic Embedding}:
Consider a text description $Y_{sem}$ of an anomalous image, for example, ``cable with black hole'' or ``grid with irregular pattern''.
We leverage the pre-trained CLIP~\cite{radford2021learning} text encoder $E_{CLIP_t}(\cdot)$ to extract features, yielding the semantic embedding $e_{sem} \in \mathcal{R}^{1 \times d}$, where $d$ represents the dimensionality of the embedding.
This process can be mathematically expressed as:
\begin{equation}
\begin{split}
\label{eq1}
e_{sem} = E_{CLIP_t}(Y_{sem}).
\end{split}
\end{equation}
To standardize the token count, the resultant embedding is replicated $k$ times.
Thereafter, we apply L2 normalization to the embedding, resulting in the final semantic embedding $e_{sem} \in \mathcal{R}^{k \times d}$.
During the training phase, the weights of the CLIP text encoder are maintained in a frozen state.

\noindent\textbf{Location Embedding}:
Given a mask $Y_{loc}$ that demarcates the location of anomalies, we employ a pre-trained mask encoder~\cite{hu2024istr}, denoted as $E_{ISTR}(\cdot)$, to extract features.
Subsequently, a one-layer convolutional neural network (CNN), represented by $E_{CNN}(\cdot)$, is utilized to adjust the dimensions, culminating in the derivation of the corresponding location embedding $e_{loc} \in \mathbb{R}^{k \times d}$.
This CNN includes convolutional layers for channel adjustments and fully connected layers to modulate the dimensions. The process can be succinctly formulated as:
\begin{equation}
\begin{split}
\label{eq2}
e_{loc} = E_{CNN}\big(E_{ISTR}(Y_{loc})\big),
\end{split}
\end{equation}
wherein the location masks are transformed into location embeddings.
Within this framework, both the mask encoder $E_{ISTR}(\cdot)$ and the CNN $E_{CNN}(\cdot)$ are subjected to fine-tuning during the training phase. This fine-tuning is essential to optimize their performance for the specific anomaly detection task.

\noindent\textbf{Textural Embedding}:
For a given image $Y_{img}$ exhibiting defects, we utilize CLIP's image encoder to extract pertinent textural features, which are then initialized within the textual embedding for subsequent training.
This procedure can be depicted by the equation:
\begin{equation}
\begin{split}
\label{eq3}
e_{tex} = E_{CLIP_i}(Y_{img}\cdot Y_{loc}).
\end{split}
\end{equation}
In this equation, a mask $Y_{loc}$ is applied to direct an attention mechanism onto the input textural information, ensuring that the extracted features are predominantly focused on the regions containing anomalies.
Following this, the textual embeddings are initialized with these features.
%
%

\noindent\textbf{X-Embedding}:
Upon acquiring embeddings from various modalities, we initially concatenate these into a unified multi-modal embedding:
\begin{equation}
\begin{split}
\label{eq4}
e_{mul} = \text{concat}(e_{sem}, e_{loc}, e_{tex}).
\end{split}
\end{equation}
Thereafter, the resulting multi-modal embedding is processed by a multi-modal aggregator.
Specifically, the aggregator employs self-attention to refine the features of the multi-modal embedding, succeeded by a multi-layer perceptron (MLP) for result remapping, with residual connections facilitating integration.
The aggregated embedding is formulated as:
\begin{equation}
\begin{split}
\label{eq5}
e_{X} = e_{X} + E_{MLP}(E_{ATT}(e_{X})),
\end{split}
\end{equation}
where $E_{MLP}(\cdot)$ is the remapping function, and $E_{ATT}(\cdot)$ symbolizes the self-attention mechanism.
The resultant embedding, denoted as the aggregated X-Embedding, serves as the final output.

\subsection{Dynamic Dif-Fusion}
\noindent\textbf{Forward Diffusion Process}:
In the forward diffusion process, we incrementally introduce noise into the images of our training dataset in a stepwise manner, leading the images to gradually deviate from their original subspace.
The distribution $q(\cdot)$ within the forward diffusion process is modeled as a Markov Chain, as delineated by the subsequent equation:
\begin{equation}
\begin{split}
\label{eq6}
q(x_1, \ldots, x_T|x_0) = \prod^T_{t=1} q(x_t|x_{t-1}),
\end{split}
\end{equation}
where $x_0\sim q(x_0)$ signifies a data point sampled from the original image distribution.
The probability density function (PDF) for the forward diffusion process is constructed as the product of individual timestep distributions ranging from $1 \to T$.
This PDF is prescribed as a Gaussian distribution, characterized by:
\begin{equation}
\begin{split}
\label{eq7}
q(x_t, |x_{t-1}) = \mathcal{N}(x_t;\sqrt{1-\beta_t}x_{t-1},\beta_t I).
\end{split}
\end{equation}
Here, $\beta$ symbolizes the precalculated diffusion rate, determined by a variance scheduler.
The identity matrix is denoted by $I$, with $\sqrt{1-\beta_t}x_{t-1}$ and $\beta_t I$ corresponding to the mean and covariance of the distribution for the image $x_t$ at timestep $t$, respectively.
The original image undergoes progressive corruption by the incremental introduction of Gaussian noise $\epsilon$ at each timestep.
By integrating Denoising Diffusion Probabilistic Models (DDPMs)~\cite{ho2020denoising} into the diffusion framework, the image at timestep $t$ can be explicitly reformulated from the initial timestep $0$ as follows:
\begin{equation}
\begin{split}
\label{eq8}
q(x_t, |x_0) = \mathcal{N}\big(x_t;\sqrt{\bar{\alpha}_t}x_{t-1},(1-\bar{\alpha}_t) I\big).
\end{split}
\end{equation}
The term $\bar{\alpha}_t$ is articulated by:
\begin{equation}
\begin{split}
\label{eq9}
\bar{\alpha}_t=\prod^t_{s=1} (1-\beta_s).
\end{split}
\end{equation}
Utilizing the reparameterization technique, $x_t$ is derived as:
\begin{equation}
\begin{split}
\label{eq10}
x_t=\sqrt{\bar{\alpha}_t}x_0+\sqrt{1-\bar{\alpha}_t}\epsilon.
\end{split}
\end{equation}
This formulation permits the sampling at any arbitrary timestep $t$ within the Markov chain.
%

  

    
    
    

\noindent\textbf{Reverse Diffusion Process}:
The objective of the reverse diffusion process is to devise a finite-time reversal strategy for the forward diffusion process, executed within a span of $T$ timesteps.
This reversal process is formulated as a Markov chain, within which a neural network is tasked with the estimation of parameters $\theta$ at each discrete timestep.
The inception of the Markov chain for the reverse diffusion occurs at the culmination point of the forward process, specifically at timestep $T$. By this juncture, the data distribution has been ostensibly transformed into an isotropic Gaussian distribution.
The process is mathematically articulated as follows:
\begin{equation}
\begin{split}
\label{eq11}
p_\theta(x_0)=\int p(x_T)\prod_{t=1}^T p_\theta(x_{t-1}|x_t),
\end{split}
\end{equation}
where the conditional distribution is defined as:
\begin{equation}
\begin{split}
\label{eq12}
p_\theta(x_{t-1}|x_t)=\mathcal{N}\big(x_{t-1};\mu_\theta(x_t,t),\Sigma_\theta(x_t,t)\big),
\end{split}
\end{equation}
where $\mu_\theta(x_t,t)$ and $\Sigma_\theta(x_t,t)$ represent the mean and covariance, respectively, as predicted on the basis of the input $x_t$ and the associated timestep $t$.

\begin{table*}[t]
\centering
\caption{ \textbf{Generation quantitative results with IS and IC-LPIPS on MVTec AD dataset}. IS and IL denote inception score and intra-cluster pairwise learned perceptual image patch similarity, respectively.}\label{tab1}
\tablestyle{0.8pt}{1.35}
\begin{tabular}{l|x{22}x{22}x{22}x{22}x{22}x{22}x{22}x{22}x{22}x{22}x{22}x{22}x{22}x{22}x{22}|x{22}}\shline
Category         & bottle & cable & caps & carp & grid & hazel & leath & metal & pill & screw & tile & brush & trans & wood & zipper & Mean \\ \hline
DiffAug~\cite{zhao2020differentiable}, IS & 1.59   & 1.72  & 1.34    & 1.19   & 1.96 & 1.67  & 2.07    & 1.58  & 1.53 & 1.10  & 1.93 & 1.33  & 1.34  & 2.05 & 1.30   & 1.58 \\
CDC~\cite{ojha2021few}, IS     & 1.52   & 1.97  & 1.37    & \textbf{1.25}   & 1.97 & 1.97  & 1.80    & 1.55  & 1.56 & 1.13  & 2.10 & 1.63  & 1.61  & 2.05 & 1.30   & 1.65 \\
Crop-P~\cite{niu2020defect}, IS  & 1.43   & 1.74  & 1.23    & 1.17   & 2.00 & 1.74  & 1.47    & 1.56  & 1.49 & 1.12  & 1.83 & 1.30  & 1.39  & 1.95 & 1.23   & 1.51 \\
SDGAN~\cite{niu2020defect}, IS   & 1.57   & 1.89  & 1.49    & 1.18   & 1.95 & 1.85  & 2.04    & 1.45  & 1.61 & 1.17  & 2.53 & 1.78  & \textbf{1.76}  & 2.12 & 1.25   & 1.71 \\
DefGAN~\cite{zhang2021defect}, IS  & 1.39   & 1.70  & 1.59    & 1.24   & 2.01 & 1.87  & \textbf{2.12}    & 1.47  & 1.61 & 1.19  & 2.35 & 1.85  & 1.47  & 2.19 & 1.25   & 1.69 \\
DFMGAN~\cite{duan2023few}, IS  & \textbf{1.62}   & 1.96  & 1.59    & 1.23   & 1.97 & 1.93  & 2.06    & 1.49  & 1.63 & 1.12  & 2.39 & \textbf{1.82}  & 1.64  & 2.12 & 1.29   & 1.72 \\
AnoDiff~\cite{hu2024anomalydiffusion}, IS & 1.58   & \textbf{2.13}  & 1.59    & 1.16   & 2.04 & 2.13  & 1.94    & 1.96  & 1.61 & 1.28  & 2.54 & 1.68  & 1.57  & \textbf{2.33} & \textbf{1.39}   & 1.80 \\
AnomalyXFusion, IS    &    \textbf{1.62}    &   \textbf{2.13}    &   \textbf{1.68}      &   1.16     &   \textbf{2.12}   &   \textbf{2.24}    & 2.00        &   \textbf{1.96}    &   \textbf{1.63}   &    \textbf{1.29}   &   \textbf{2.55}   &   1.80    &   1.64    &   2.09   &    1.37    &  \textbf{1.82}    \\ \hline\hline
DiffAug~\cite{zhao2020differentiable}, IL & 0.03   & 0.07  & 0.03    & 0.06   & 0.06 & 0.05  & 0.06    & 0.29  & 0.05 & 0.10  & 0.09 & 0.06  & 0.05  & 0.30 & 0.05   & 0.09 \\
CDC~\cite{ojha2021few}, IL     & 0.04   & 0.19  & 0.06    & 0.03   & 0.07 & 0.05  & 0.07    & 0.04  & 0.06 & 0.11  & 0.12 & 0.06  & 0.13  & 0.03 & 0.05   & 0.07 \\
Crop-P~\cite{niu2020defect}, IL  & 0.04   & 0.25  & 0.05    & 0.11   & 0.12 & 0.21  & 0.14    & 0.15  & 0.11 & 0.16  & 0.20 & 0.08  & 0.15  & 0.23 & 0.11   & 0.14 \\
SDGAN~\cite{niu2020defect}, IL   & 0.06   & 0.19  & 0.03    & 0.11   & 0.10 & 0.16  & 0.12    & 0.28  & 0.07 & 0.10  & 0.21 & 0.03  & 0.13  & 0.25 & 0.10   & 0.13 \\
DefGAN~\cite{zhang2021defect}, IL  & 0.07   & 0.22  & 0.04    & 0.12   & 0.12 & 0.19  & 0.14    & 0.30  & 0.10 & 0.12  & 0.22 & 0.03  & 0.13  & 0.29 & 0.10   & 0.15 \\
DFMGAN~\cite{duan2023few}, IL  & 0.12   & 0.25  & 0.11    & 0.13   & 0.13 & 0.24  & 0.17    & 0.32  & 0.16 & 0.14  & 0.22 & 0.18  & 0.25  & 0.35 & 0.27   & 0.20 \\
AnoDiff~\cite{hu2024anomalydiffusion}, IL & \textbf{0.19}   & 0.41  & \textbf{0.21}    & \textbf{0.24}   & 0.44 & 0.31  & 0.41    & 0.30  & 0.26 & \textbf{0.30}  & \textbf{0.55} & 0.21  & 0.34  & 0.37 & 0.25   & 0.32 \\
AnomalyXFusion, IL    &    \textbf{0.19}    &   \textbf{0.42}    &    0.20     &   0.22     &   \textbf{0.45}   &   \textbf{0.33}    &     \textbf{0.42 }   &    \textbf{0.34}   &   \textbf{0.27}   &   0.29    &  0.52    &    \textbf{0.22}   &     \textbf{0.36}  &   \textbf{0.38}   &  \textbf{ 0.26}     &   \textbf{0.33}  \\ \shline
\end{tabular} 
\end{table*}
\begin{table*}[t]
\centering
\caption{ \textbf{Generation quantitative results with IS and IC-LPIPS on MVTec LOCO dataset}. IS and IL denote inception score and intra-cluster pairwise learned perceptual image patch similarity, respectively.}\label{tab2}
\tablestyle{3.0pt}{1.35}
\begin{tabular}{l|cc|cc|cc|cc|cc|c} \shline
\multirow{2}{*}{Category} & \multicolumn{2}{c|}{breakfast box} & \multicolumn{2}{c|}{juice bottle} & \multicolumn{2}{c|}{pushpins} & \multicolumn{2}{c|}{screw bag} & \multicolumn{2}{c|}{splicing connectors} & \multicolumn{1}{c}{\multirow{2}{*}{Mean}} \\ 
 & \multicolumn{1}{c}{logical} & \multicolumn{1}{c|}{structural} & \multicolumn{1}{c}{logical} & \multicolumn{1}{c|}{structural} & \multicolumn{1}{c}{logical} & \multicolumn{1}{c|}{structural} & \multicolumn{1}{c}{logical} & \multicolumn{1}{c|}{structural} & \multicolumn{1}{c}{logical} & \multicolumn{1}{c|}{structural} & \multicolumn{1}{c}{} \\ \hline
DFMGAN~\cite{duan2023few}, IS & 1.21 & 1.33 & 1.12 & 1.17 & 1.02 & 1.14 & 1.34 & 1.49 & 1.18 & 1.72 & 1.27 \\
AnoDiff~\cite{hu2024anomalydiffusion}, IS & 1.28 & 1.49 & 1.21 & 1.25 & 1.08 & 1.31 & 1.67 & 1.73 & 1.35 & 1.98 & 1.44 \\
AnomalyXFusion, IS & \textbf{1.39} & \textbf{1.60} & \textbf{1.36} & \textbf{1.43} & \textbf{1.18} & \textbf{1.52} & \textbf{2.04} & \textbf{2.37} & \textbf{1.64} & \textbf{2.41} & \textbf{1.69} \\ \hline \hline
DFMGAN~\cite{duan2023few}, IL & 0.24 & 0.23 & 0.13 & 0.13 & 0.23 & 0.24 & 0.20 & 0.21 & 0.33 & 0.23 & 0.22 \\
AnoDiff~\cite{hu2024anomalydiffusion}, IL & 0.29 & 0.27 & 0.18 & 0.16 & 0.37 & 0.30 & 0.26 & 0.25 & 0.40 & 0.27 & 0.28 \\
AnomalyXFusion, IL & \textbf{0.35} & \textbf{0.30} & \textbf{0.24} & \textbf{0.18} & \textbf{0.46} & \textbf{0.32} & \textbf{0.35} & \textbf{0.31} & \textbf{0.47} & \textbf{0.34} & \textbf{0.33} \\ \shline
\end{tabular} 
\end{table*}

\noindent\textbf{Dynamic Diffusion Process}:
To achieve optimal integration of multi-modal information at each timestep, we aim to adaptively adjust the X-embedding $e_X$.
This approach ensures that the weighting of different embeddings is balanced across various timesteps.
Consequently, we define the following conditional probability function:
\begin{equation}
\begin{split}
\label{eq13}
p_\theta(x_0)=\int p(x_T)\prod_{t=1}^T p_\theta(x_{t-1}|x_t,e_t),
\end{split}
\end{equation}
where $e_t$ is defined as a function of the timestep:
\begin{equation}
\begin{split}
\label{eq14}
e_t=f_\phi(e_X,t),
\end{split}
\end{equation}
with $\phi$ representing the learnable parameters of the function.
In our implementation, a conditioned MLP is utilized to realize this dynamic adjustment.
Specifically, the conditioned MLP aligns the timestep to the same dimensionality as $e_X$ to produce $e_{tim}$, which is then concatenated and fed into the MLP.
Subsequently, during the reverse diffusion process, $e_{t}$ is employed to determine the mean and covariance, as expressed by:
\begin{equation}
\begin{split}
\label{eq15}
x_{t-1}=\mu_\theta(x_t,e_t)+\sqrt{\Sigma_\theta(x_t,e_t)}\epsilon,
\end{split}
\end{equation}
Cross-attention mechanisms are employed to perform these operations.
Drawing inspiration from the blended diffusion approach presented in~\cite{avrahami2022blended}, we integrate both the input image and the corresponding mask into the training process to enhance the focus on the anomaly region, yielding:
\begin{equation}
\begin{split}
\label{eq16}
x_{t-1}^{'}&=y_{loc}\cdot\big(\mu_\theta(x_t,e_t)+\sqrt{\Sigma_\theta(x_t,e_t)}\epsilon\big)\\
&+(1-y_{loc})\cdot\big(\sqrt{\bar{\alpha}_t}x_0+\sqrt{1-\bar{\alpha}_t}\epsilon\big),
\end{split}
\end{equation}
Finally, our loss function is defined as:
\begin{equation}
\begin{split}
\label{eq17}
\mathcal{L}=\mathbb{E}_{t,x_0,\epsilon}\big[\| \epsilon-\epsilon_\theta(x_t,e_t,t) \|\big].
\end{split}
\end{equation}
Thereafter, the AnomalyXFusion framework can be trained holistically using Eq.~\ref{eq17}.

\begin{figure*}
\centering
\includegraphics[width=0.98\linewidth]{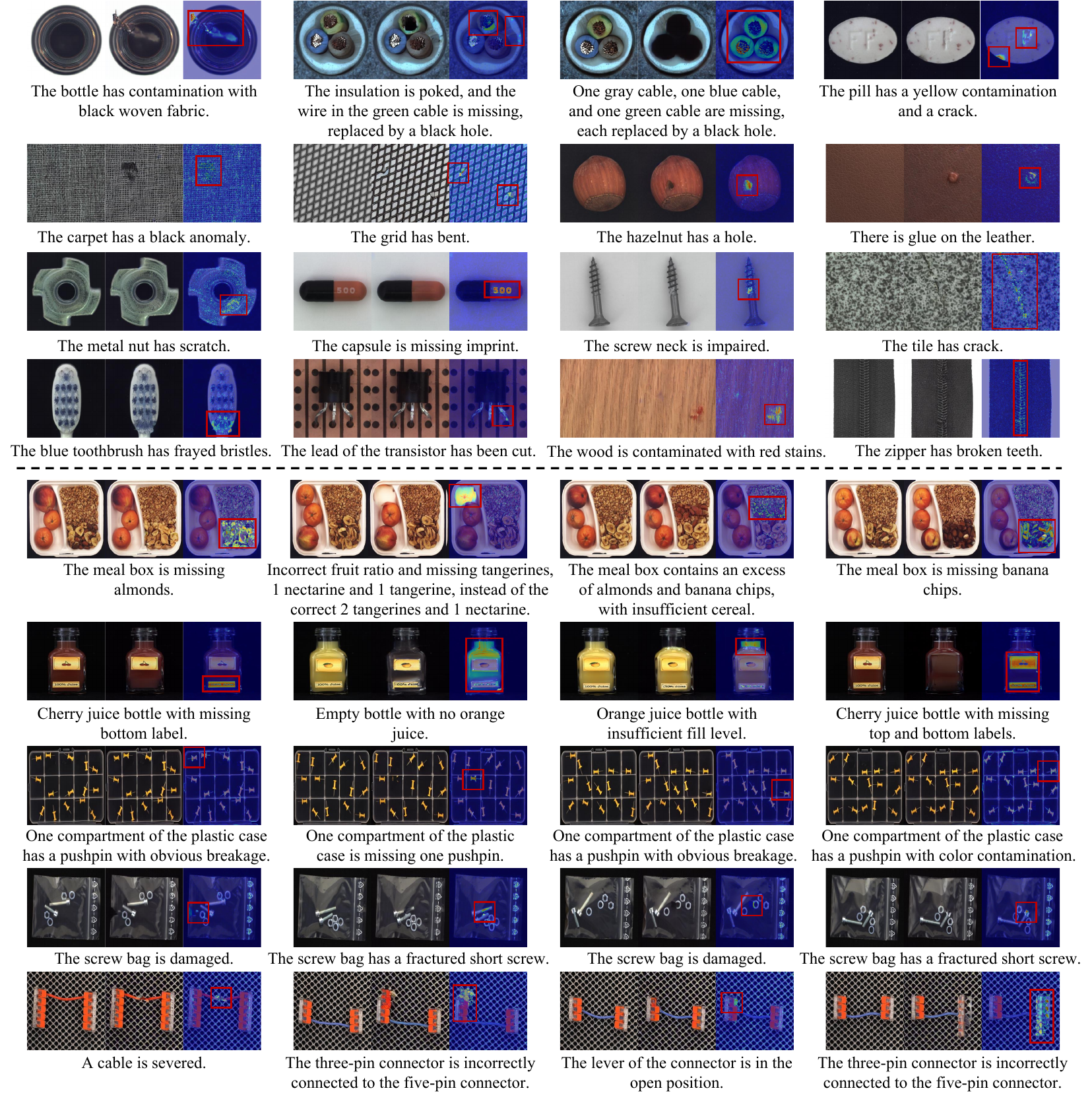} 
\vspace{-2.0mm}
\caption{
\textbf{Qualitative Evaluation.} The qualitative assessment is facilitated by visualizing the output images, which are organized in sets from left to right as follows: the original anomaly-free image, the generated image with induced defects, the heatmap indicating the location of the defect generation, and the corresponding textual description.
}\label{fig4}
\end{figure*}

\begin{table*}[t]
\centering
\caption{\textbf{Results of anomaly detection on MVTec AD dataset}.
In the table, AUC-P, AP-P, and F$_1$-P denote the results of anomaly localization (pixel-level).
AUC-I, AP-I, and F$_1$-I denote the results of anomaly detection (image-level).
}\label{tab3}
\tablestyle{0.3pt}{1.35}
\begin{tabular}{l|x{22}x{22}x{22}x{22}x{22}x{22}x{22}x{22}x{22}x{22}x{22}x{22}x{22}x{22}x{22}|x{22}}\shline
Category         & bottle & cable & caps & carp & grid & hazel & leath & metal & pill & screw & tile & brush & trans & wood & zipper & Mean \\ \hline
DREAM~\cite{zavrtanik2021draem}, AUC-P     & 96.7   & 80.3  & 76.2    & 92.6   & 99.1 & 98.8  & 98.5    & 96.9  & 95.8 & 91.0  & 98.5 & 93.8  & 76.5  & 98.8 & 93.4   & 92.2 \\
PRN~\cite{zhang2023prototypical}, AUC-P       & 97.5   & 94.5  & 95.6    & 96.4   & 98.9 & 98.0  & 99.4    & 97.9  & 98.3 & 94.0  & 98.5 & 96.1  & 94.9  & 96.2 & 98.4   & 96.9 \\
DFMGAN~\cite{duan2023few}, AUC-P    & 98.9   & 97.2  & 79.2    & 90.6   & 75.2 & 99.7  & 98.5    & 99.3  & 81.2 & 58.8  & 99.5 & 96.4  & 96.2  & 95.3 & 92.9   & 90.0 \\
AnoDiff~\cite{hu2024anomalydiffusion}, AUC-P   & 99.4   & \textbf{99.2}  & \textbf{98.8}    & 98.6   & 98.3 & \textbf{99.8}  & \textbf{99.8}    & \textbf{99.8}  & \textbf{99.8} & 97.0  & 99.2 & 99.2  & 99.3  & \textbf{98.9} & \textbf{99.4}   & 99.1 \\
AnomalyXFusion, AUC-P      &   \textbf{99.6}     &    97.3   &    98.4     &  \textbf{99.1}      &   \textbf{100}   &     99.5  &    \textbf{99.8}     &   \textbf{99.8}    & 99.7     &    \textbf{98.1}   &   \textbf{99.9}   &   \textbf{99.6}    &    \textbf{99.9}   &  98.6    &    \textbf{99.4}    &    \textbf{99.3}  \\ \hline\hline
DREAM~\cite{zavrtanik2021draem}, AP-P      & 80.2   & 21.8  & 25.5    & 43.0   & 59.3 & 73.6  & 67.6    & 84.2  & 45.3 & 30.1  & 93.2 & 29.5  & 31.7  & 87.8 & 65.4   & 54.1 \\
PRN~\cite{zhang2023prototypical}, AP-P        & 76.4   & 64.4  & 45.7    & 69.6   & 58.6 & 73.9  & 58.1    & 93.0  & 55.5 & 47.7  & 91.8 & 46.4  & 68.6  & 74.2 & 79.0   & 66.2 \\
DFMGAN~\cite{duan2023few}, AP-P     & 90.2   & 81.0  & 26.0    & 33.4   & 14.3 & 95.2  & 68.7    & 98.1  & 67.8 & 2.2   & 97.1 & 75.9  & 81.2  & 70.7 & 65.6   & 62.7 \\
AnoDiff~\cite{hu2024anomalydiffusion}, AP-P    & 94.1   & \textbf{90.8}  & 57.2    & 81.2   & 52.9 & \textbf{96.5}  & 79.6    & 98.7  & \textbf{97.0} & \textbf{51.8}  & 93.9 & 76.5  & 92.6  & 84.6 & 86.0   & 81.4 \\
AnomalyXFusion, AP-P       &   \textbf{96.7}     &   79.8    &   \textbf{57.6}      &  \textbf{84.6}      &   \textbf{97.3}   &    90.9   &  \textbf{91.7}       &  \textbf{99.0}     &  96.1    &   30.4    &   \textbf{99.0}   &   \textbf{90.4}    &   \textbf{98.5}    &  \textbf{92.2}    &    \textbf{87.8}    &  \textbf{86.1}    \\ \hline\hline
DREAM~\cite{zavrtanik2021draem}, F$_1$-P   & 74.0   & 28.3  & 32.1    & 41.9   & 58.7 & 68.5  & 65.0    & 74.5  & 53.0 & 35.7  & 87.8 & 28.4  & 24.2  & 80.9 & 64.7   & 53.1 \\
PRN~\cite{zhang2023prototypical}, F$_1$-P     & 71.3   & 61.0  & 47.9    & 65.6   & 58.9 & 68.2  & 54.0    & 87.1  & 72.6 & 49.8  & 84.4 & 46.2  & 68.4  & 67.4 & 73.7   & 64.7 \\
DFMGAN~\cite{duan2023few}, F$_1$-P  & 83.9   & 75.4  & 35.0    & 38.1   & 20.5 & 89.5  & 66.7    & 94.5  & 72.6 & 5.3   & 91.6 & 72.6  & 77.0  & 65.8 & 64.9   & 62.1 \\
AnoDiff~\cite{hu2024anomalydiffusion}, F$_1$-P & 87.3   & \textbf{83.5}  & \textbf{59.8}    & 74.6   & 54.6 & \textbf{90.6}  & 71.0    & 94.0  & \textbf{90.8} & \textbf{50.9}  & 86.2 & 73.4  & 85.7  & 74.5 & \textbf{79.2}   & 76.3 \\
AnomalyXFusion, F$_1$-P    &   \textbf{90.5}     &   73.2    &  53.1       &    \textbf{76.6}    &   \textbf{91.0}   &    82.8   &     \textbf{82.6}    &   \textbf{95.0}    &   89.0   &   38.3    &   \textbf{94.6}   &   \textbf{84.0}    &   \textbf{93.1}    &   \textbf{85.3}   &   \textbf{79.2}     &  \textbf{80.6}    \\ \hline\hline
DREAM~\cite{zavrtanik2021draem}, AUC-I     & 99.3   & 72.1  & 93.2    & 95.3   & 99.8 & 100   & 100     & 97.8  & 94.4 & 88.5  & 100  & 99.4  & 79.6  & 100  & 100    & 94.6 \\
PRN~\cite{zhang2023prototypical}, AUC-I       & 94.9   & 86.3  & 84.9    & 92.6   & 96.6 & 93.6  & 99.1    & 97.8  & 88.8 & 84.1  & 91.1 & 100   & 88.2  & 77.5 & 98.7   & 91.6 \\
DFMGAN~\cite{duan2023few}, AUC-I    & 99.3   & 95.9  & 92.8    & 67.9   & 73.0 & 99.9  & 99.9    & 99.3  & 68.7 & 22.3  & 100  & 100   & 90.8  & 98.4 & 99.7   & 87.2 \\
AnoDiff~\cite{hu2024anomalydiffusion}, AUC-I   & 99.8   & \textbf{100}   & \textbf{99.7}    & 96.7   & 98.4 & \textbf{99.8}  & \textbf{100}     & \textbf{100}   & 98.0 & 96.8  & \textbf{100}  & \textbf{100}   & \textbf{100}   & 98.4 & 99.9   & \textbf{99.2} \\
AnomalyXFusion, AUC-I      &   \textbf{100}     &   99.7    &     97.2    &    \textbf{96.8}    &    \textbf{100}  &  \textbf{99.8}     &     \textbf{100}    &   \textbf{100}    &   \textbf{98.8}   &    \textbf{97.2}   &  \textbf{100}    &  \textbf{100}     &   \textbf{100}    &   \textbf{98.5}   &   \textbf{100}     &   \textbf{99.2}   \\ \hline\hline
DREAM~\cite{zavrtanik2021draem}, AP-I      & 99.8   & 83.2  & 98.7    & 98.7   & 99.9 & 100   & 100     & 99.6  & 98.9 & 96.3  & 100  & 99.8  & 80.5  & 100  & 100    & 97.0 \\
PRN~\cite{zhang2023prototypical}, AP-I        & 98.4   & 92.0  & 95.8    & 97.8   & 98.9 & 96.0  & 99.7    & 99.5  & 97.8 & 94.7  & 96.9 & 100   & 88.9  & 92.7 & 99.7   & 96.6 \\
DFMGAN~\cite{duan2023few}, AP-I     & 99.8   & 97.8  & 98.5    & 87.9   & 90.4 & 100   & 100     & 99.8  & 91.7 & 64.7  & 100  & 100   & 92.5  & \textbf{99.4} & 99.9   & 94.8 \\
AnoDiff~\cite{hu2024anomalydiffusion}, AP-I    & 99.9   & \textbf{100}   & \textbf{99.9}    & \textbf{98.8}   & 99.5 & \textbf{99.9}  & \textbf{100}     & \textbf{100}   & 99.6 & 97.9  & \textbf{100}  & \textbf{100}   & \textbf{100}   & 99.4 & \textbf{100}    & 99.7 \\
AnomalyXFusion, AP-I       &    \textbf{100}    &    99.8   &    \textbf{99.9}     &    \textbf{98.8 }   &   \textbf{100}   &  99.8     &    \textbf{100}     &     \textbf{100}  &   \textbf{99.9}   &   \textbf{99.6}    &  \textbf{100}    &  \textbf{100}     &   \textbf{100}    &  99.1    &   \textbf{100}     &  \textbf{99.8}    \\ \hline\hline
DREAM~\cite{zavrtanik2021draem}, F$_1$-I   & 98.9   & 79.2  & 94.0    & 93.4   & 98.8 & 100   & 100     & 97.6  & 95.8 & 89.3  & 100  & 97.6  & 71.4  & 100  & 100    & 94.4 \\
PRN~\cite{zhang2023prototypical}, F$_1$-I     & 94.1   & 84.0  & 94.3    & 92.1   & 95.0 & 94.1  & 97.6    & 96.9  & 93.2 & 87.2  & 89.3 & 100   & 84.0  & 86.7 & 97.6   & 92.4 \\
DFMGAN~\cite{duan2023few}, F$_1$-I  & 97.7   & 93.8  & 94.5    & 87.3   & 85.4 & 99.0  & 99.2    & 99.2  & 91.4 & 85.3  & 100  & 100   & 88.9  & 98.8 & 99.4   & 94.7 \\
AnoDiff~\cite{hu2024anomalydiffusion}, F$_1$-I & 98.9   & \textbf{100}   & \textbf{98.7}    & 94.3   & 98.7 & 98.9  & \textbf{100}     & \textbf{100}   & 97.0 & \textbf{95.5}  & \textbf{100}  & \textbf{100}   & \textbf{100}   & \textbf{98.8} & 99.4   & \textbf{98.7} \\
AnomalyXFusion, F$_1$-I    &    \textbf{100}    &    98.3   &    98.5     &    \textbf{96.2}    &   \textbf{100}   &   \textbf{98.2}    &    \textbf{100}     &   99.2    &   \textbf{97.9}   &    93.0   &   \textbf{100}   &  \textbf{100}     &   \textbf{100}    &    98.7  &    \textbf{100}    &   \textbf{98.7} \\ \shline
\end{tabular} 
\end{table*}

\section{Experiments}

\subsection{Experimental Settings}
\noindent\textbf{Datasets}:
The MVTec AD dataset~\cite{bergmann2021mvtec,bergmann2019mvtec} consists of over 5,000 high-resolution images, spanning fifteen distinct classes of objects and textures. Each class is represented by a set of defect-free training images and a test set that includes images with various defects, as well as images devoid of defects. The dataset also offers pixel-level annotations for all the anomalies present.
The MVTec LOCO dataset~\cite{bergmann2022beyond} includes both structural and logical anomalies, with a total of 3,644 images distributed across five categories. Structural anomalies are characterized by physical imperfections such as scratches, dents, or contaminations on manufactured products. Logical anomalies, on the other hand, indicate a breach of inherent constraints, exemplified by the incorrect placement of permissible objects or the lack of required objects. This dataset also provides pixel-level ground truth data for each anomalous region.
In this study, we introduce an enhancement to the MVTec AD and LOCO datasets through the addition of annotated captions, designated as MVTec AD Caption and LOCO Caption, respectively. These captions have been carefully designed to ensure multimodal complementarity, significantly enhancing the ability to identify and distinguish logical anomalies.

\noindent\textbf{Evaluation Metrics}:
We employ the Inception Score (IS), a metric that is independent of the provided anomaly data, to assess the quality of the generated anomalies. Furthermore, we introduce the Intra-cluster Pairwise Learned Perceptual Image Patch Similarity (IC-LPIPS) distance metric to quantify the diversity of the synthesized anomalies.
The Frechet Inception Distance (FID) and Kernel Inception Distance (KID) are not utilized in our assessment due to the limited availability of anomaly data.
For the evaluation of anomaly perception, we utilize the Area Under the Receiver Operating Characteristic curve (AUROC), Average Precision (AP), and the F$_1$-max score to measure the precision and effectiveness of anomaly detection and localization tasks.

\subsection{Results on Anomaly Generation}
To substantiate the efficacy of our generative model, we have selected DiffAug~\cite{zhao2020differentiable}, CDC~\cite{ojha2021few}, Crop-P~\cite{niu2020defect}, SDGAN~\cite{niu2020defect}, DefGAN~\cite{zhang2021defect}, DFMGAN~\cite{duan2023few}, and AnoDiff~\cite{hu2024anomalydiffusion} as comparative benchmarks.
The quantitative results of the generation task on the MVTec AD dataset are presented in Table~\ref{tab1}.
We synthesized 1,000 images per category to enable the calculation of numerical metrics.
Upon analysis, it is evident that GAN-based methods generally surpass traditional techniques, with diffusion-based models achieving even higher performance compared to GAN-based approaches.
When juxtaposed with the state-of-the-art anomaly generation method, AnoDiff~\cite{hu2024anomalydiffusion}, our proposed approach has demonstrated a more holistic enhancement.
For instance, in image quality metrics, our approach obtained a score of 1.82, which is slightly higher than AnoDiff's score of 1.80.
Furthermore, our method has demonstrated enhanced diversity, particularly in categories such as metal nuts and zippers, where notable improvements were observed.
The quantitative results for the generation task on the MVTec LOCO dataset are detailed in Table~\ref{tab2}.
An evaluation of these results indicates that DFMGAN and AnoDiff struggle to produce satisfactory outcomes for images containing logical and structural anomalies.
In contrast, our method has provided commendable results, suggesting a higher proficiency in managing complex anomaly patterns. This positions our approach as a promising alternative for generating logical and structural anomalies within the industrial imaging sector.

\begin{table*}[t]
\centering
\caption{ \textbf{Results of anomaly classification on MVTec AD dataset}. }\label{tab4}
\tablestyle{2.0pt}{1.35}
\begin{tabular}{l|x{22}x{22}x{22}x{22}x{22}x{22}x{22}x{22}x{22}x{22}x{22}x{22}x{22}x{22}|x{22}}\shline
Category         & bottle & cable & caps & carp & grid & hazel & leath & metal & pill & screw & tile & trans & wood & zipper & Mean \\ \hline
DiffAug~\cite{zhao2020differentiable}          & 48.84  & 21.36 & 34.67   & 35.48  & 28.33 & 65.28 & 40.74   & 58.85 & 29.86 & 25.10 & 59.65 & 38.09 & 41.27 & 22.76  & 39.31 \\
CDC~\cite{ojha2021few}              & 38.76  & 39.06 & 28.89   & 25.27  & 35.83 & 54.86 & 43.38   & 48.44 & 21.88 & 32.92 & 48.54 & 29.76 & 28.57 & 14.63  & 35.06 \\
Crop-P~\cite{niu2020defect}           & 52.71  & 32.81 & 32.89   & 27.96  & 28.33 & 59.03 & 34.39   & 59.89 & 26.74 & 28.81 & 68.42 & 41.67 & 47.62 & 26.42  & 40.55 \\
SDGAN~\cite{niu2020defect}            & 48.84  & 21.88 & 30.22   & 21.50  & 30.83 & 43.75 & 38.10   & 44.27 & 20.49 & 26.75 & 42.69 & 32.14 & 30.95 & 21.54  & 32.43 \\
DefGAN~\cite{zhang2021defect}           & 53.49  & 21.36 & 32.00   & 29.03  & 27.50 & 61.11 & 42.33   & 56.77 & 28.47 & 28.81 & 26.90 & 35.72 & 24.60 & 18.70  & 34.77 \\
DFMGAN~\cite{duan2023few}           & 56.59  & 45.31 & 37.23   & 47.31  & 40.83 & 81.94 & 49.73   & 64.58 & 29.52 & 37.45 & 74.85 & 52.38 & 49.21 & 27.64  & 49.61 \\
AnoDiff~\cite{hu2024anomalydiffusion}          & 90.70  & 67.19 & 66.67   & 58.06  & 42.50 & 85.42 & 61.90   & 59.38 & 59.38 & 48.15 & 84.21 & 60.71 & 71.43 & 69.51  & 66.09 \\
AnomalyXFusion             &   \textbf{93.35}     &   \textbf{71.00}    &    \textbf{71.33}     &  \textbf{68.97}      &    \textbf{69.32}   &    \textbf{89.75}   &    \textbf{67.78}     &    \textbf{66.88}   &   \textbf{67.50}    &   \textbf{50.27}    &    \textbf{89.01}   &   \textbf{76.43}    &    \textbf{82.71}   &    \textbf{81.46}    &   \textbf{74.70}    \\ \shline
\end{tabular} 
\end{table*}

\begin{table*}[t]
\centering
\caption{ \textbf{Results of anomaly classification on MVTec LOCO dataset}.}\label{tab5}
\tablestyle{3.6pt}{1.35}
\begin{tabular}{l|cc|cc|cc|cc|cc|c} \shline
\multirow{2}{*}{Category} & \multicolumn{2}{c|}{breakfast box} & \multicolumn{2}{c|}{juice bottle} & \multicolumn{2}{c|}{pushpins} & \multicolumn{2}{c|}{screw bag} & \multicolumn{2}{c|}{splicing connectors} & \multicolumn{1}{c}{\multirow{2}{*}{Mean}} \\ 
 & \multicolumn{1}{c}{logical} & \multicolumn{1}{c|}{structural} & \multicolumn{1}{c}{logical} & \multicolumn{1}{c|}{structural} & \multicolumn{1}{c}{logical} & \multicolumn{1}{c|}{structural} & \multicolumn{1}{c}{logical} & \multicolumn{1}{c|}{structural} & \multicolumn{1}{c}{logical} & \multicolumn{1}{c|}{structural} & \multicolumn{1}{c}{} \\ \hline
DFMGAN~\cite{duan2023few} & 60.46 & 40.34 & 54.23 & 40.67 & 14.56 & 30.67 & 29.63 & 1.34 & 29.42 & 58.34 & 35.97 \\
AnoDiff~\cite{hu2024anomalydiffusion} & 70.12 & 47.16 & 64.33 & 53.23 & 21.98 & 38.12 & 41.08 & 5.98 & 40.00 & 72.35 & 45.44 \\
AnomalyXFusion & \textbf{76.00} & \textbf{50.00} & \textbf{72.00} & \textbf{60.34} & \textbf{29.00} & \textbf{40.00} & \textbf{47.00} & \textbf{6.38} & \textbf{46.00} & \textbf{82.76} & \textbf{50.95} \\ \shline
\end{tabular} 
\end{table*}

Additionally, the visualization results are depicted in Fig.~\ref{fig4}.
The provided examples clearly demonstrate that our methodology is proficient in generating a diverse array of high-quality defect data based on various textual prompts.
This adaptability highlights the robustness of our model in interpreting a wide range of textual cues, which is instrumental in crafting a more comprehensive and nuanced dataset.
The diversity observed in the generated defect imagery is indicative of the model's advanced comprehension and interpretation capabilities regarding textual prompts. This enables the model to produce a broad spectrum of defect variations that are closely aligned with the specific textual cues provided.
This capability is particularly advantageous for applications that require the recognition and resolution of a multitude of defect types, thereby augmenting the model's applicability in real-world industrial settings for defect detection and diagnosis.

\subsection{Results on Anomaly Perception}
\noindent\textbf{Anomaly Localization}:
We have undertaken a comparative analysis of our approach against existing anomaly generation methods, focusing on their performance in anomaly localization.
For each method, we generated 1,000 images per anomaly type using the respective models and subsequently trained corresponding U-Net architectures~\cite{ronneberger2015u} for comparative evaluation.
The localization outcomes were aggregated using average pooling to derive confidence scores for image-level anomaly detection, a technique aligned with the approach adopted by DREAM~\cite{zavrtanik2021draem}.
Table~\ref{tab3} presents the comparative results of anomaly localization, as measured by AUC-P, AP-P, and F$_1$-P.
The data within the table clearly demonstrate that our method outperforms previous approaches in overall performance and has shown marked improvements in areas that were previously challenging.
For example, within the grid defects category, we have achieved an AP-P value of 97.3, representing a substantial enhancement over previous methods.

\begin{table}[t]
\centering
\caption{ \textbf{Ablation study on MVTec dataset}. In the table, MIF denotes Multi-modal In-Fusion, and DDF denotes Dynamic Dif-Fusion.}\label{tab6}
\tablestyle{9.4pt}{1.35}
\begin{tabular}{ccc|c|c|c} \shline
Text & MIF & DDF & IS & IL & Acc \\ \hline
  -   &   -     &    -     & 1.63  & 0.23 & 65.38  \\
$\surd$    &    -    &     -    &  1.65 & 0.26 & 67.39 \\
$\surd$    & $\surd$      &    -     & 1.73 & 0.29 & 69.47  \\
$\surd$    & -      &    $\surd$     & 1.76 & 0.28 & 69.54  \\\hline
$\surd$    & $\surd$      & $\surd$       & \textbf{1.82} & \textbf{0.33} & \textbf{74.70}  \\ \shline
\end{tabular} 
\end{table}

\begin{figure}
\centering
\includegraphics[width=1.0\linewidth]{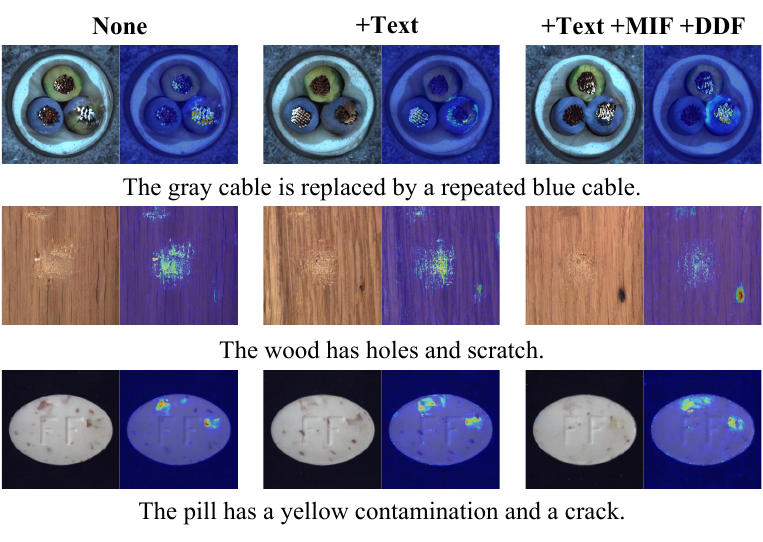} 
\vspace{-5.5mm}
\caption{
\textbf{Visualization of component impacts}. The results reveal that without the incorporation of text information, the attributes of the generated defects could not be precisely controlled. Upon introducing text cues, the generated defects began to reflect the text input, yet the quality of the defects remained suboptimal. However, with the integration of all components, we achieved the generation of high-quality defects with a high degree of controllability.
}\label{fig5}
\end{figure}

\noindent\textbf{Anomaly Detection}:
The outcomes of the anomaly detection task are articulated in Table~\ref{tab3}, where performance is measured using metrics AUC-I, AP-I, and F$_1$-I.
An examination of the table reveals that our methodology has yielded commendable outcomes, demonstrating comparable results to the existing SOTA methods in the domain of anomaly detection.
This comparative analysis affirms the competence of our approach in anomaly detection when benchmarked against established techniques.

\noindent\textbf{Anomaly Classification}:
To substantiate the effectiveness of the anomaly generation models, a ResNet-34 classifier was employed for performance evaluation.
Consistent with the methodologies of AnoDiff and DFMGAN, the generated data were used for training, followed by validation on a separate test set.
For the MVTec AD dataset, the results, as detailed in Table~\ref{tab4}, show that our approach achieved a classification accuracy of 74.70, which notably exceeds the performance of previous methodologies.
Specifically, our method surpassed the current SOTA AnoDiff by 8.61 points.
In analogous fashion, for the MVTec LOCO dataset, comparative analysis with DFMGAN and AnoDiff was conducted. The data presented in Table~\ref{tab5} demonstrate that our method also outperformed these methods, thereby reinforcing the robustness and dependability of our anomaly generation model across various datasets.

\subsection{Ablation Study}
Our ablation studies have systematically evaluated the contribution of several pivotal components within our proposed framework to their effectiveness.
Specifically, we have assessed the efficacy of the Text Condition, Multi-modal In-Fusion (MIF), and Dynamic Diffusion (DDF) modules through both the generation task and the subsequent anomaly classification task.
The outcomes of these studies are delineated in Table~\ref{tab6}.
Analysis of the tabulated data reveals that the absence of these modules resulted in generated metrics of 1.63 and 0.23, with a classification accuracy capped at 65.38.
The incorporation of the Text Condition module led to a discernible enhancement in classification accuracy, increasing to 67.39.
Further augmentation of accuracy was realized with the introduction of the MIF and DDF modules, culminating in the highest precision rates observed in our experiments.
Supporting these quantitative results, Fig.~\ref{fig5} visually demonstrates the influence of the respective modules on the generation process.
Visual assessment indicates a diminished control over the generation outcomes in the absence of these modules.
While the Text Condition alone did not produce significant improvements, which we surmise may be due to inadequate modality integration, the sequential addition of the MIF and DDF modules substantially enhanced the controllability and quality of the generated anomalies, concomitantly elevating classification performance.

\section{Conclusion}
We present the AnomalyXFusion framework, designed to mitigate the issue of limited abnormal sample availability in the context of industrial manufacturing, with a particular focus on intricate structural and logical anomalies. This framework, characterized by its dual-module architecture, amalgamates multi-modal data through the Multi-modal In-Fusion (MIF) module and exerts control over the anomaly generation process via the Dynamic Dif-Fusion (DDF) module. Furthermore, we introduce the MVTec Caption dataset, which comprises 2.2k annotated image-text pairs, enhancing the precision of anomaly description.
Empirical assessments have demonstrated that AnomalyXFusion surpasses contemporary methodologies in the authenticity and diversity of anomaly generation, consequently enhancing the efficacy of anomaly detection procedures. The framework's adeptness in generating complex anomalies is a notable advancement in the domain. The broader implications of this research are significant, laying the groundwork for subsequent studies in multi-modal anomaly synthesis and underscoring the critical role of multi-faceted data integration in the evolution of anomaly detection methodologies.

\backmatter

\bmhead{Data Availability}
The MVTec Caption dataset is publicly available at \url{https://github.com/hujiecpp/MVTec-Caption}
\bmhead{Code Availability}
The codes for the described experiments are available at \url{https://github.com/hujiecpp/MVTec-Caption}
\bibliographystyle{sn-mathphys}
\bibliography{sn-bibliography}

\end{document}